# ListReader: Extracting List-form Answer for Opinion Questions

**Peng Cui**, **Dongyao Hu**, and **Le Hu**

Harbin Institute of Technology, Harbin, China

{pcui, lhu, dyhu}@insun.hit.edu.cn

## Abstract

Question answering (QA) is a high-level ability of natural language processing. Most extractive machine reading comprehension models focus on factoid questions (e.g., who, when, where) and restrict the output answer as a short and continuous span in the original passage. However, in real-world scenarios, many questions are non-factoid (e.g., how, why) and their answers are organized in the list format that contains multiple non-contiguous spans. Naturally, existing extractive models are by design unable to answer such questions. To address this issue, this paper proposes *ListReader*, a neural extractive QA model for list-form answer. In addition to learning the alignment between the question and content, we introduce a heterogeneous graph neural network to explicitly capture the associations among candidate segments. Moreover, our model adopts a co-extraction setting that can extract either span- or sentence-level answers, allowing better applicability. Two large-scale datasets of different languages are constructed to support this study. Experimental results show that our model considerably outperforms various strong baselines. Further discussions provide an intuitive understanding of how our model works and where the performance gain comes from. Code and data will be made public[1].

## 1 Introduction

Recent years have seen a surge of interest on extractive question answering system. Machine reading comprehension (MRC) is one representative task which aims to read one or a set of passages and then answer the questions about it. Existing MRC models usually formulate it as an extractive task where the answer is a text span from the input passage. With the development of large-scale datasets, such as CNN/Daily Mail (Hermann et al. 2015), SQuAD (Rajpurkar et al. 2016; Rajpurkar, Jia, and Liang 2018), and NarrativeQA (Kocisky et al. 2018), a lot of reading comprehension models (Seo et al. 2017; Yu et al. 2018) have achieved promising results. Some of them even surpass human performance (Yang et al., 2019, Lan et al., 2020).

---

[1]We will release our code and data after the publication of this paper.

Figure 1. An open question with its list-form answer. The complete answer consists of multiple non-contiguous spans and each answer contains richer semantic than the entity-type answer. Such characteristics present a challenge for existing extractive MRC models.

Despite their success, we found there are two main limitations that have not been addressed by previous studies. First, most extractive models assume the target answer is a short and continuous span in the given passage, which could be rather restrictive for real-life applications because many questions require a list-form answer that contains multiple non-contiguous sentences or spans. However, those models cannot be applied in such situation effectively since the "*boundary prediction*" paradigm is not suitable for recognizing multiple answers (Hu et al., 2019; Segal et al., 2020).

Second, previous studies mainly focus on factoid questions such as "*when*", "*where*", and "*who*". Since the answers of these questions are limited to certain types of words such as

"*time*", "*place*" and "*person name*", the desired reasoning process ability sometimes will degenerate to entity recognizing. For example, empirical observations in (Sugawara et al., 2018) show that only using the first several words of the questions can also achieve a good result. By contrast, answering opinion questions, such as "*how*" and "*why*", has been rarely explored. Because of the open-ended nature of such questions, their answers usually have more flexible formats and richer semantics than a single-span and entity-like answer, thus requiring deeper text understanding (Jia and Liang, 2017).

To illustrate the abovementioned challenges, we take an example in Figure 1, where the passage lists three answers for the question. Such form of QA is prevalent in real-life scenarios, whereas it is a challenge for existing methods. Without understanding the relation between "*cola*" between "*rust Remover*", models are likely to miss the second answer "*apply cola to the surface*" because it does not have explicit semantic associations with "*screw*". However, such inter-answer reasoning ability is usually neglected by previous models. Besides, how to extract multiple discontinuous answers is a challenge.

In this paper, we propose a novel extractive QA model for answering opinioned questions with list-form answers. We first use a BERT (Devlin et al., 2018) encoder to learn the contextual representations of the question and passage. Then, we adopt a heterogeneous graph neural network to simultaneously capture the $\langle question, passage \rangle$ alignment and inter-answer dependencies. Moreover, we unify the span- and sentence-level information in one framework, enabling the joint promotion of each other. To evaluate our model, we collected two datasets covering various domains of questions, such as sport, health, etc. Experimental results demonstrate the effectiveness and superiority of our model. To summarize, our contributions are three-fold:

- We study a practical but underexplored QA task where the question is opinion based and the answer contains multiple spans. We construct two datasets. Both of them consists of high-quality list-form QA pairs covering diverse domains and therefore are the ideal test-beds for our model and future studies of similar tasks.

- We propose the *ListReader*, a neural extractive QA that introduces the inter-answer relations with a graph neural network for multi-answer extraction.

- Experimental results on the constructed datasets demonstrate that our approach considerably outperforms various strong baseline methods. Further ablation study validates the power of the model components.

## 2 Model

This section describes the proposed model *ListReader*, of which Figure 2 presents an overall architecture. Let $Q = \{q_1, ..., q_m\}$ and $P = \{p_1, ..., p_n\}$ be the input question and passage, respectively. In general, the target answer list of our task can be formally defined as $A = \{a_1, a_2, ..., a_t\}$, where $a_i$ is either a span- or sentence-level segment from $P$. In our task, $t \geq 2$ and different answers are non-contiguous. The goal of our model is to extract every answer of A.

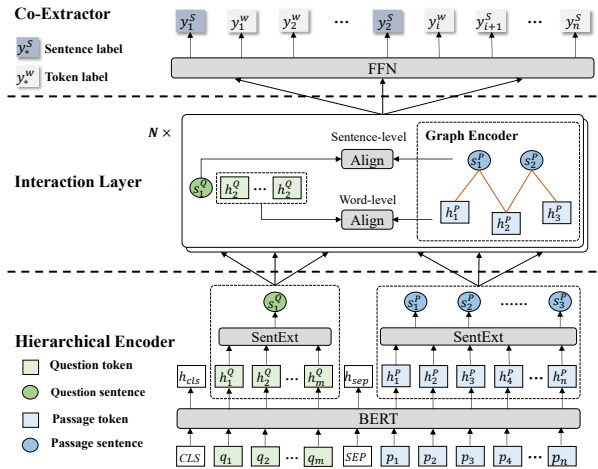

Figure 2. Main architecture of our model, which consists of three components: 1) The **hierarchical encoder** (bottom) generates token and sentence representations of the question and passage; 2) The **interaction layer** (middle) captures the $\langle Question, Answer \rangle$ alignment and inter-answer relations simultaneously; 3) The **co-extractor** (top) jointly predicts answer spans and the sentences they appear.

### 2.1 Hierarchical Encoder

The hierarchical encoder aims to learn the contextual information of the $Q$ and $P$ for further extraction. It adopts a hierarchical structure that first obtains token representations, which are then aggregated to generate sentence representations.

**Token Encoding** Based on the current development of NLP (Radford et al. 2019; Vaswani et al., 2017), we encode the question and passage sequences with a BERT encoder, which has shown promising results in extractive QA task (Lan et al., 2020). Similar to previous study, we concatenate the question and passage into a flat sequence, where a $[CLS]$ token and a $[SEP]$ token are inserted to the beginning and end of the question, respectively. Then, we send the whole sequence to the BERT encoder.

$$\mathbf{H^Q}, \mathbf{H^P} = \text{BERT}([CLS], q_1, ..., q_m, [SEP], p_1, ..., p_n), \quad (1)$$

where $\mathbf{H^Q} = \{h_1^Q, h_2^Q, ..., h_m^Q\}$ and $\mathbf{H^P} = \{h_1^P, h_2^P, ..., h_n^P\}$ are the token representations of $Q$ and $P$, respectively.

**Sentence Encoding** Since BERT can only output token- and document-level representations, we add a self-attentive sentence extractor (SentExt) on the top of BERT to obtain sentence-level representations. Let $\{h_{k,1}^P, h_{k,2}^P, ..., h_{k,c}^P\}$ be the hidden states of the $k$-th sentence learned by BERT. The SentExt module can be denoted as:

$$a_i^k = \mathbf{W_2} \cdot \tanh(\mathbf{W_1} \cdot h_{k,i}^P + \mathbf{b_1}) + \mathbf{b_2}, \quad (2)$$

$$\alpha_i^k = \frac{\exp\left(a_i^k\right)}{\sum_{j=1}^c \exp\left(a_j^k\right)}, \tag{3}$$

$$s_k^p = \sum_{i=1}^c \alpha_i^k \cdot h_{k,i}^p, \tag{4}$$

where $\alpha_i^k$ is the normalized score of $i$-th token in $k$-th sentence. $s_k^p$ is the weighted sum of all tokens in sentence. In this paper, $\mathbf{W}_*$ and $\mathbf{b}_*$ are trainable parameters for all equations.

We use $\mathbf{S}^Q = \{s_1^Q, s_2^Q, ..., s_k^Q\}$ and $\mathbf{S}^P = \{s_1^Q, s_2^Q, ..., s_l^Q\}$ to denote the extracted sentence representations of the question and passage, which are then feed to the interaction layer together with word representations $\mathbf{H}^Q, \mathbf{H}^P$.

## 2.2 Interaction Layer

As discussed in Section 1, extracting a list-form answer not only requires to model the relevance between the question and passage content, but also to capture the relations across different answers. Therefore, we construct the interaction layer with two modules: (1) The *align module* employs the attention mechanism to build the interaction between the question and passage in both token- and sentence-level, which is a standard step in almost every previous QA models (Yu et al., 2018). (2) The *graph encoder* represents the passage as a heterogeneous graph and encode it with a graph convolution network (GCN; Kipf and Welling 2017), allowing the inter-answer reasoning.

**Question-Passage Alignment** Similar to previous studies, the alignment calculation consists of two parts, which are *question-to-passage* alignment and *passage-to-question* alignment. Specifically, we first obtain a word similarity matrix $\mathbf{U}_w \in \mathbb{R}^{m \times n}$, where $u_{i,j}$ represents the relevance between $i$-th word in $\boldsymbol{P}$ and $j$-th word in $\boldsymbol{Q}$. A heuristic function is applied to to compute $u_{i,j}$, as shown as follows:

$$u_{i,j} = \mathbf{W}_3 \cdot (h_i^Q, h_j^P, |h_i^Q - h_j^P|, h_i^Q \odot h_j^P, \mathbf{b}_3), \tag{5}$$

where $h_*^*$ is the word representation learned by BERT encoder (Eq. 1). $\odot$ denotes the point-wise operation.

Afterwards, we normalized $\mathbf{U}_w$ by row, rendering a matrix $\overline{\mathbf{U}}_w$. The passage-to-question alignment is obtained through $\mathbf{H}^{P\prime} = \overline{\mathbf{U}}_w^\top \cdot \mathbf{H}^Q$. As for the question-to-passage alignment, we have $\mathbf{H}^{Q\prime} = \widetilde{\mathbf{U}}_w \mathbf{U}_w^\top \cdot \mathbf{H}^P$, where matrix $\widetilde{\mathbf{U}}_w$ is obtained by normalizing each column of $\mathbf{U}_w$.

The sentence-level alignment can be achieved in a similar way, where the sentence similarity matrix $\mathbf{U}_s \in \mathbb{R}^{k \times l}$ with $\mathbf{S}^Q$ and $\mathbf{S}^P$ learned from the SentExt module (Eq.4).

**Inter-Answer Reasoning** The effect of inter-answer reasoning has been rarely explored by previous studies because the alignment module already provides enough information for single-span answer extraction. However, such feature plays a critical role for the multi-answer task. To address this issue, we introduce a graph encoder to build the inter-answer interaction upon the aligned passage representations.

Our passage graph is constructed as follows. Given an input passage, we use the passage words and sentences to build a heterogeneous graph. The edge set $V$ is defined as $V = V_s \cup V_w$, where $V_s$ contains $l$ sentence nodes and $V_s$ contains $m$ word nodes. The edge set can be formulated as $E = \{e_{1,1}, ..., e_{i,j}, ..., e_{l,m}\}$, where $e_{ij}$ is a real-value and $e_{ij} > 0$ indicates that $i$-th sentence contains $j$-th word.

Given the constructed graph, we sent it into a stacked graph encoder. Let $G^{(i)} \in \mathbb{R}^{(l+m) \times d}$ be the node representations in the $i$-th layer, the updating process can be denoted as:

$$G^{(i+1)\prime} = \text{Relu}(\widetilde{A} \cdot \text{Relu}(\widetilde{A} \cdot G^{(i)} \mathbf{W}_4 + \mathbf{b}_4) \mathbf{W}_5 + \mathbf{b}_5), \tag{6}$$

$$G^{(i+1)} = \text{LayerNorm}(G^{(i)} + G^{(i+1)\prime}), \tag{7}$$

where Eq.6 is a two-layer GCN. $\widetilde{A} = D^{-1/2} A D^{-1/2}$ is the normalized symmetric adjacency matrix. D is the degree matrix, and A is the adjacency matrix derived from TF-IDF features. Eq.7 is a residual connection followed with a Layer Normalization. Here, we initialize $G^{(0)}$ with the word and sentence representations learned from the alignment module, which informs the graph encoder which segments are important to the question and should be focused on.

In our experiments, multiple interaction layers are stacked and therefore the $\langle$Question, Answer$\rangle$ alignment and inter-answer reasoning are performed iteratively, enabling the joint consideration of the two features in list-form answer extraction. For notation simplicity, we still use $\mathbf{H}^Q, \mathbf{H}^P, \mathbf{S}^Q$, and $\mathbf{S}^P$ to denote the aligned word and sentence representations of the question and passage.

## 2.3 Co-Extractor

We adopt a co-extraction setting that joint predicts the token- and sentence-level labels. We do this for the following two reasons. First of all, a span-level answer could be too concise to answer an opinion question, whereas the whole sentence where the answer span lies in may provide detailed explanations for better understanding. Second, empirical observations (Segal et al., 2019) have proved that relevant content retrieval is a critical step for extractive MRC task. As a results, the sentence selection and span extraction could play complementary roles and promote the performance of each other (Yuan et al., 2020).

**Span Extraction** Most extractive MRC models adopt the "*boundary prediction*" strategy that locates the answer span by predicting its begin and end token. However, our target answers contain multiple spans and their lengths might vary widely because the questions are opinion based. Inspired by the well-known `BIO` tagging approach, we regard the multi-answer extraction task as a sequence labelling task in which `B` corresponds to the begin token, `I` corresponds to the inside token, and `O` corresponds to the out-of-answer token. For the $i$-th token in $P$ with its representation $h_i^P$, we compute its label $y_w^i$ using a feed-forward layer, i.e.

$$y_w^i = \text{softmax}(\mathbf{W}_6 h_i^P + \mathbf{b}_6), \tag{8}$$

**Sentence Prediction** For the $i$-th sentence $s_i^P$, we compute its label through $\hat{y}^i = \text{softmax}(\mathbf{W}_7 s_i^P + \mathbf{b}_7)$, where $\hat{y}^i \in \{0,1\}$ indicates whether it is an answer sentence. In our experiments, we employ a multi-task learning strategy to jointly

train the span extraction and sentence prediction task. The final loss of our model is expressed as:

$$\mathcal{L} = \mathcal{L}_w + \lambda \mathcal{L}_s, \tag{9}$$

Where $\mathcal{L}_w$ and $\mathcal{L}_s$ represent the cross-entropy loss of span extraction and sentence selection, respectively. $\lambda$ is the hyperparameter to balance the two parts.

## 3 Experimental Setup

### 3.1 Datasets

Most existing public datasets are constructed for factoid and single-span answer extraction, thus not suitable for our task. To evaluate our model and encourage future studies, we construct two QA datasets: WikiHowQA and WebQA. The statistics of the two datasets are summarized in Table 1. Their detailed description and collection process are described as follows.

| Datasets | Splits | | | # Ans. |
|---|---|---|---|---|
| | Train | Val. | Test | |
| WikiHowQA | 197,936 | 25,000 | 25,000 | 7.4 |
| WebQA | 31,597 | 4,000 | 4,000 | 4.9 |

Table 1. Statistics of two datasets. "Ans." denotes the average number of answer span for each question.

**WikiHowQA** is an English dataset collected from www.wikihow.com, a hot QA website where the questions and answers are all carefully edited by human. Most pages of this site are in a specific pattern where the title is a "*How to*" question and the content contains a list-type answer. Therefore, we can automatically obtain the ⟨Question, AnswerList⟩ pairs by parsing the html structure. The whole dataset is randomly split to 197,936/25,000/25,000 for train, validation, and test. Although all questions of this dataset begin with the fixed words "*how to*", we found that the questions cover diverse topics, such as sports, health, education, etc.

**WebQA** is a Chinese dataset collected from structural web pages. We first fetch a large number of raw html pages from about 90 human-selected QA sites. Then, we select the pages according to the following criteria: (1) The page title is a question, which can be recognized by question keywords and patterns. (2) The answers scattered in different positions of the passage, which can be recognized by rules and patterns based on the page structures. Finally, we check the selected pages by human and obtain the final dataset, including 31,597/4,000/4,000 for train, validation, and test.

### 3.2 Implementation Detail

We use the pre-trained "bert-base-uncased" version BERT with hidden size of 768. We stacked three interaction layer in our experiments. In each layer, the graph encoder consists of a two-layer GCN followed with a layer normalization with highway connection. The output dimension of interaction layer remains 768. In the joint prediction, since the number sentence is far less than the words, we set $\lambda = 2$ (i.e. the weight of sentence selection loss) to prevent the collapse of this part.

| Models | WikiHowQA | | WebQA | |
|---|---|---|---|---|
| | Span | Sent. | Span | Sent. |
| BERT-base | 60.7 | 65.1 | 60.3 | 63.5 |
| QANet | 61.4 | 70.3 | 59.4 | 64.7 |
| TASE | 65.6 | 68.3 | 61.8 | 66.0 |
| MTMSN | 64.4 | 67.9 | 64.9 | 68.3 |
| ListReader | **72.3** | **76.9** | **73.6** | **75.8** |

Table 2. F1 score of different models. *Span* and *Sent.* indicate the span-level and sentence-level extraction performance, respectively.

We train our model with 4 NVIDIA P100 cards for up to 1000 epoch, and the batch size is set to 16. During the training, we use Adam to optimize the parameters with a learning rate of $1e-4$. An early-s top strategy is applied when the validation loss no longer declines for 10 batch. The experimental results reported in this paper are averaged over 5 runs.

### 3.3 Models for Comparison

We compare our model with the following representative extractive QA models:

- **BERT-base** (Devlin et al., 2019)   BERT is a bidirectional transformer encoder pre-trained with large-scale corpus. We send the question and passage to the BERT and directly use the BERT outputs for prediction.
- **QANet** (Zhao et al., 2017)   is a high performance extractive QA model. We modify the training objective to test whether normal single-span extraction model can be generalized to multiple answer extraction.
- **MSN** (Hu et al., 2019)   Multi-span network addresses the multi-span extraction problem in two steps. First, it introducing an additional module to predict the answer number K. Then, a beam search is applied to pick K answers.
- **TASE** (Segal et al., 2020)   Tag based span extraction applies BIO tagging to multi-span MRC task. However, it is designed for factoid questions and ignores the inter-answer reasoning.

## 4 Results and Analysis

This section describes the experimental results. We first compare our model with strong baseline models. Following previous studies, we use F1 score as the evaluation metric for both span or sentence level extraction[2]. Then, we conduct an ablation and case study to show where the performance gain comes from and how our model works.

### 4.1 Main Results

We can obtain the following observations from Table 1: 1) Our model beats all baseline models on both span and sentence level prediction, proving the effectiveness of our model.

---

[2]We add the SentExt module (Eq. 2-4) to the baseline models to extract sentence representations for sentence-level extraction.

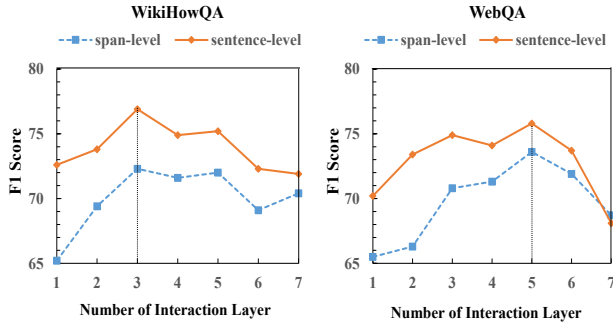

Figure 4. Relations between the Number of Interaction Layer $N$ and model performance on the two datasets.

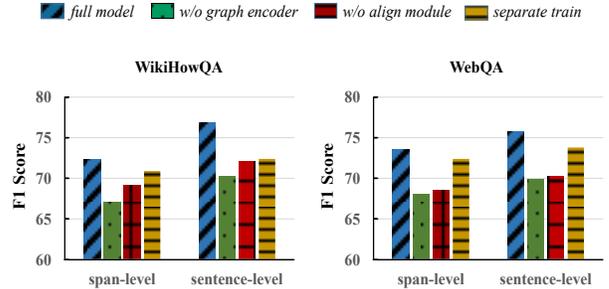

Figure 3. Results comparison between our full model and its three ablated variants on two datasets.

2) The two strong baselines BERT-base and QANet fail to achieve the desired results on our datasets compared to their success on single-span MRC task. Such result shows that normal single-span MRC models cannot generalize to the multi-answer extraction scenario effectively, which is consistent with previous observation. 3) Although MSN and TASE are particularly designed for multi-span answer extraction, our model outperforms them by a large margin on the two datasets. One possible reason is that they are built for factoid questions whose answers are entity words without rich semantic associations among them. As a result, the inter-answer reasoning ability is neglected, leading to the poor performance on answering opinion questions.

**Influence of the Degree of Interaction**  Recall in our experiments, we stacked multiple interaction layer to combine the question-passage alignment and inter-answer association features. To analyze, we conduct a quantitative analysis on the influence of the number of interaction layer. As can be seen from Figure 4, results on different tasks and datasets show a similar trend. When N tunes from a small value, the performance considerably increases. However, such boost reaches a saturation when N exceeds a threshold, which is 3 and 5 for WikiHowQA and WebQA datasets, respectively.

Noticeably, the F1 score of span-level extraction is highly consistent with that of sentence-level extraction on both datasets, indicating the two tasks are indeed related. Nevertheless, we find that when N is set to a small value, the performance span-level extraction shows a more significant decline than that of sentence-level extraction. This observation indicates that span-level extraction requires deeper interaction than sentence-level extraction.

### 4.3 Ablation Study

To probe into the relative contributions of different modules and understand where the performance gain comes from, we construct three ablated variants for comparison.

• **w/o graph encoder**  removes the inter-answer reasoning module (i.e. graph encoder). This module is usually neglected by previous single-span extraction model. We use this variant to test whether it is necessary for our task.

• **w/o align module**  removes the align module of interaction layer and is used to analyze whether the matching features can be learned in the BERT encoder.

• **separate train**  trains the sentence prediction and span extraction task independently, which can be regard as two models. Although observations in Figure 3 has shown the consistency between main task and auxiliary task, we use this variant to obtain a quantitative analysis of the effect of this setting.

As shown in Figure 4, Our full model outperforms all ablated versions, proving that each component is necessary and combing them can help our model achieve the best results. Noticeably, the performance declines dramatically when removing either the graph encoder or alignment module. This observation implies that the two modules play Complementary roles, which will be further analyzed in Section 4.4. We also note that the introduction sentence prediction task can boost the performance gain on span-level extraction, consistent with the results of Yuan et al. (2020).

### 4.4 Analysis of Interaction Layer

In this subsection, we present a case study to better understand how our model works on extracting list-form answer. To this end, we send the hidden states of each sublayer of the interaction layer to the prediction layer and observe which sentences are selected in the current layer. In this way, we can have trace the decision process of our model.

Figure 5 presents visualization result of an instance from WikiHowQA dataset. From the results, we can have several intuitive observations. In the 1-th layer, the alignment module roughly selects the question-related sentences, while missing the 5-th answer sentence since "" have no explicit semantic associations. However, the graph encoder recalls this answer in the following learning because it captures the relation between "*cola*" and "*rust remover*", which is the keyword of the first answer (1-th sentence). Such process shows the power of inter-answer reasoning ability for multi-answer extraction.

We also note that the graph encoder could introduce noisy sentences. For example, it mistakes sentence 3, 11, and 15 in the first and second layer (shown in $G^{(1)}$ and $G^{(2)}$). Under this circumstance, the following alignment module correctly filter these sentences (shown in $A^{(2)}$ and $G^{(3)}$). This observation implies that the two modules play complementary roles, consistent with the results in Figure 4.

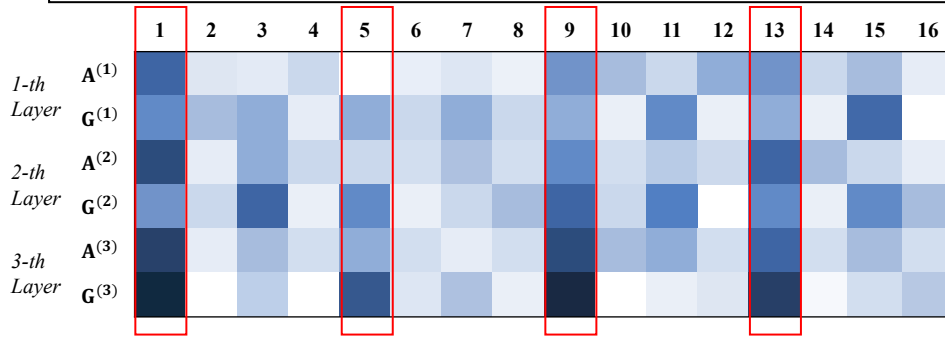

**Question**: *How to remove rusted screws?*

**Paragraph**: [1] Soak the screw in rust remover for 15 minutes. [2] Commercial rust remover often comes in a spray bottle, so all you need to do is point the nozzle and shoot. [3] Spray a lot of penetrant around the screw's head. [4] This should lubricate the head as well as allow penetrant to drip down into the screw's shaft. [5] Apply the cola to the surface. [6] If you don't have rust remover at home, you can use cola instead. [7] This is because that cola contains carbonic acid, which can react with iron oxide, the main composition of rust. [8] Other drinks can achieve a similar effect. [9] Strike the screw a few times with a metal hammer. [10] Square up the hammer so it is directly over the screw's head. [11] Rapidly hit the head a few times to break the rust seal holding the screw in place. [12] Use some force if you can, enough to jar the screw while still maintaining your accuracy. [13] Loosening Screws with Heat. [14] Cleaning the screw is especially important after attempting to remove it through other methods. [15] Heat can cause rust penetrant and other chemicals to catch fire. [16] To prevent this, dampen a rag with the degreaser and wipe down the screw quickly. …

Figure 5. A visualized case study to illustrate the inner decision process of the interaction layer. For space limitation, we select a paragraph that contains four answers and 16 sentences. The heat map shows the probability of each sentence in different layers. Darker cell indicates higher probability. $A^{(i)}$ and $G^{(i)}$ denote the alignment module and graph encoder in the $i$-th interaction layer (described in Section 3.2), respectively.

These observations to some extend explain the effectiveness of our model. The alignment module selects relevant spans as candidates, which provides a seed set for the graph encoder to further search implicit answer throughout inter-answer reasoning. In the other hand, the alignment module can filter noisy results, ensuring the correctness.

## 5 Related Work

Machine reading comprehension is an important high-level natural language application. According to the answer type, existing MRC models and datasets can be roughly divided into three types: 1) **multiple-choices**, 2) **cloze**, and 3) **extraction**. Early studies mainly focused on first two forms. In multiple-choice task, a question is combined with four options. Datasets of this setting includes MCTest (Richardson, Burges, and Renshaw 2013), Bioprocess (Berant et al. 2014), and Race (Lai et al. 2017). Cloze is another type of MRC which removes key words of contents and asks the model to fill the bank based on the context. Hermann et al., (2015) blank out the entities of CNN/DM mails. Children's Book Test (CBT; Hill et al., 2016) removes random entities and gives the previous 20 sentences as context.

Recent studies focus more and more on extractive MRC, which aims to extract one or several spans from original document as answers, e.g. the classical dataset SQuAD (Rajpurkar et al. 2016) and BiDAF model (Seo et al., 2016). In more realistic scenarios, however, the answer is often composed by or inferred from several items, e.g. dataset DROP (Dua et al.,

2019) exposes the shortcomings of single-span extractive method. In recent years, few studies have focus on the multispan answer extraction task. Hu et al., (2019) proposed MTMSN model. It first predicts the required number of answers, then extracts multiple spans to specific amount. In comparison, Segal et al. (2020) proposes TASE (TAg-based Span Extraction) which adopts the well-known BIO tagging scheme to MRC and converts it to a sequence tagging task. Nevertheless, these approaches are mostly built based on the factoid QA datasets where the answer is an entity-like span, which would reduce the expected reasoning ability to entity recognition (Jia and Liang et al. 2017). Besides, most of them focus on short answers with several words, whereas the long form answer extraction has been rarely explored.

## 6 Conclusion and Future Work

This study focus on an underexplored QA situation where the answer consists of multiple discontinuous spans and the question is opinion based. We propose the *ListReader* to address this issue, a neural extractive QA model that highlights the inter-answer reasoning and is able to extract both span- and sentence-level list-form answers. Two automatic constructed and human revised datasets are collected to support this study. Experimental results demonstrate that our model outperforms various strong baselines. Further analysis validates the effectiveness of our modules. In the future, we will introduce knowledge base to help answer the questions of different domains.